%% file: main.tex
\documentclass[runningheads]{llncs}

\usepackage{eccv}

\usepackage{eccvabbrv}

\usepackage{graphicx}
\usepackage{booktabs}
\usepackage{multirow}
\usepackage{tabularx}
\usepackage{adjustbox}
\usepackage{amssymb}
\usepackage{pifont}
\newcommand{\cmark}{\ding{51}}%
\newcommand{\xmark}{\ding{55}}%
\usepackage{makecell} 
\usepackage[accsupp]{axessibility}  

\usepackage{bibunits}

\defaultbibliographystyle{splncs04}
\defaultbibliography{main}

\usepackage{hyperref}

\usepackage{orcidlink}
\newcommand{\dataname}{Event\-Kitchen\xspace}

\begin{document}

\title{Cooking beyond Frames: A Stereo Event Camera Dataset in the Kitchen} 

\titlerunning{Cooking beyond Frames: A Stereo Event Camera Dataset in the Kitchen}

\author{Chengming Feng\inst{1}\orcidlink{0009-0008-0314-4735} \and
Hesam Araghi\inst{1}\orcidlink{0000-0002-4539-4408} \and
Liming Zheng\inst{1}\orcidlink{0000-0002-7544-3020} \and
Julien Dupeyroux\inst{2}\orcidlink{0000-0002-7414-5021} \and
Xucong Zhang\inst{1}\orcidlink{0000-0002-8368-3542} \and
Jan van Gemert\inst{1}\orcidlink{0000-0002-3913-2786} \and
Nergis Tömen\inst{1}\orcidlink{0000-0003-3916-1859}
}

\authorrunning{C.Feng et al.}

\institute{Delft University of Technology, The Netherlands \\ \email{\{c.feng-1, n.tomen\}}@tudelft.nl \and
STMicroelectronics, France
}

\maketitle

\begin{bibunit}[splncs04]
\input{sec/0_abstract}    
\input{sec/1_intro}

\input{sec/2_relatedwork}

\input{sec/3_dataset}

\input{sec/4_experiments}
\input{sec/5_conclusion}
\section*{Acknowledgements}
We thank Ruud de Jong from INSY, TU Delft, for technical support with the hardware, and Qingru Li from ImPhys, TU Delft, for assistance with the optical lenses and filters.

%
%
\putbib[main]
\end{bibunit}

\clearpage
\begin{bibunit}[splncs04]
\input{sec/suppl}
\putbib[main]
\end{bibunit}

\end{document}

%% file: sec/0_abstract.tex
\begin{abstract}
Event cameras, also known as neuromorphic cameras, have gained significant attention in recent years due to their high temporal resolution, high dynamic range, and low power consumption. While many studies and datasets in neuromorphic vision have focused on automotive and drone applications, human-centric daily-life scenarios remain largely underrepresented, despite their importance for developing and benchmarking event-based perception systems. Moreover, the few existing event-based human activity datasets are typically recorded with scripted human actions, limiting their ability to capture natural human behaviors.
In this paper, we introduce \textbf{\dataname}, a large-scale stereo event camera benchmark dataset of human cooking activities in the kitchen. \dataname is egocentrically collected from 10 participants in 13 diverse kitchens, where the participants wear a helmet with multiple sensors and naturally perform cooking activities, without any scripted actions. \dataname comprises 5.5 hours of stereo event recordings with synchronized RGB, depth, and IMU data. We provide human annotations for 10,762 action segments and 13,482 bounding boxes. We train baseline models on \dataname to perform multiple event-based tasks, including action recognition, object detection, and stereo depth estimation. By capturing natural, real-world human activities, \dataname establishes a challenging benchmark for neuromorphic vision beyond autonomous driving. The dataset and toolkit are available at \url{https://chengmingf.github.io/EventKitchen.github.io/}
\end{abstract}

%% file: sec/1_intro.tex
\section{Introduction}
\label{sec:introduction}

Unlike standard frame-based RGB cameras, event cameras operate asynchronously, where each pixel independently responds to intensity changes. This sensing paradigm of event cameras provides extremely high temporal resolution, high dynamic range, and low power consumption~\cite{eventreview}. As a result, event cameras excel at capturing rapid object movements with reduced motion blur, and preserving fine object features under challenging illumination conditions, significantly outperforming traditional frame-based sensors in dynamic and high-contrast scenarios~\cite{eventreview,eventnature,timelens,deblur,deblur2}. 
Due to these advantages, event cameras are popular in autonomous driving~\cite{ddd20,gen1,eventnature,dsec,m3ed,sevd,mvsec,eventDriveReview}. This, in turn, led to the development of several large-scale, real-world datasets, such as 1Mpx~\cite{1mpx} and GEN1~\cite{gen1}, which target event-based object detection, as well as multi-task datasets like DSEC~\cite{dsec} and MVSEC~\cite{mvsec}. In addition, the low-cost computation makes event cameras favorable in robotics and drone research~\cite{eventdroneGuido, eventdroneUZH, eventdroneUZH2, eventdroneUZH3, m3ed, mvsec}.

\begin{figure*}[!t]
    \centering
  \includegraphics[width=0.98\textwidth]{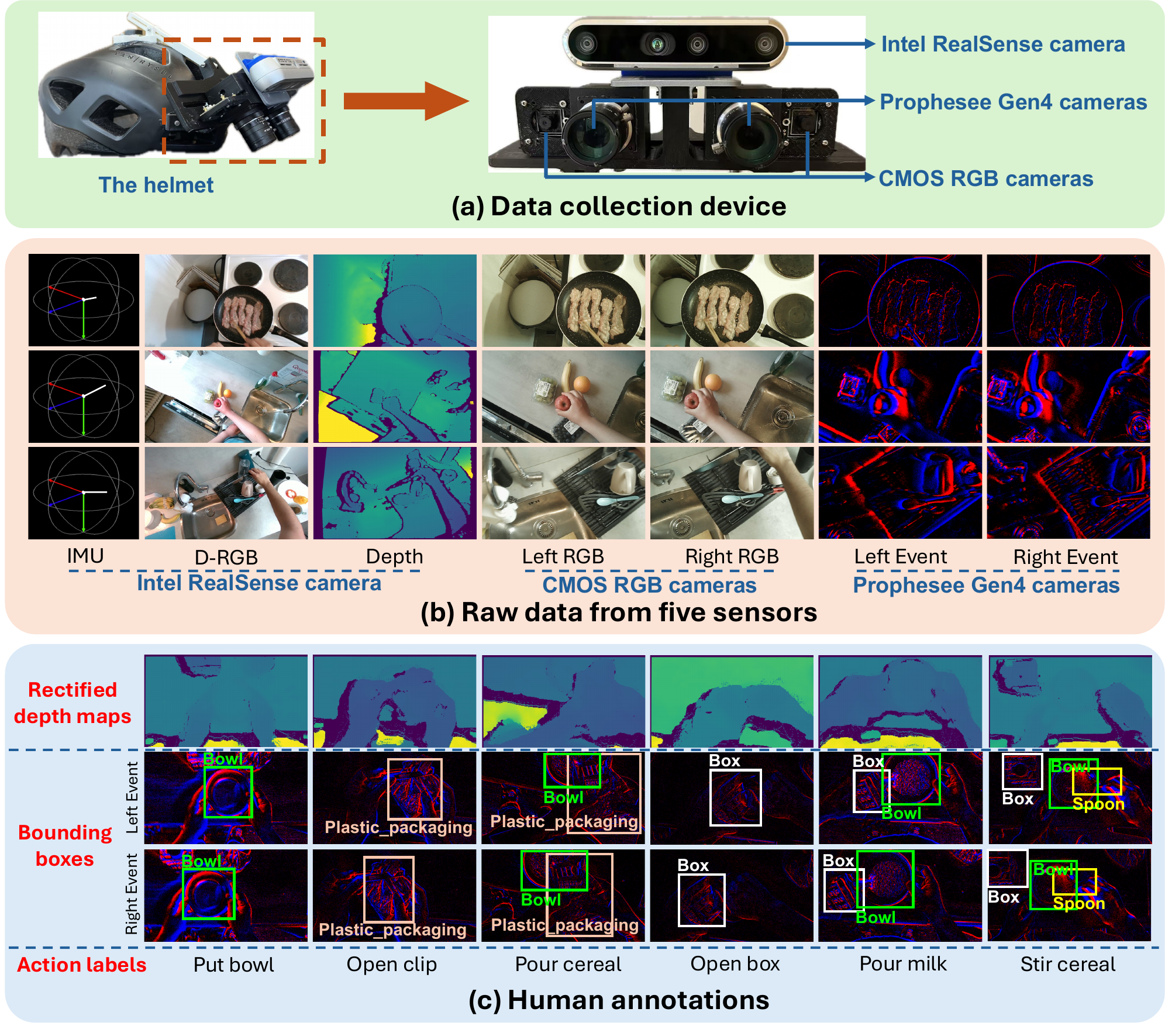}
    \caption{(a) \textbf{Data collection device}: The device is a wearable bicycle helmet with mounted sensors including two event cameras, two CMOS RGB cameras, and one Intel RealSense depth camera to allow multi-modal sensing; (b) \textbf{Raw data}: We show seven example raw data streams collected with the device in three out of 13 diverse kitchens, columns `IMU', \mbox{`D-RGB'}, and `Depth' are from Intel RealSense, `Left RGB' and `Right RGB' from two CMOS stereo cameras, `Left Event' and `Right Event' from stereo event cameras; (c) \textbf{Human annotations}: We show example human annotations including action labels and bounding boxes, which enables the tasks of event-based action recognition and object detection. The collected depth maps can be rectified to both left and right event cameras to support the event-based stereo depth estimation. 
    }
    \label{fig:teaser_figure}
\end{figure*}

In contrast, the study of event-based human daily-life activities remains significantly underexplored, despite their importance for developing and benchmarking neuromorphic perception systems in human-centric environments~\cite{land}.
Although event cameras have demonstrated advantages in capturing fine-grained and fast human motion for action recognition~\cite{dvsgesture,hardvs,nepickkitchen,spikeAction,seact} and pose estimation~\cite{EE3D-R,pose2,pose3}, existing datasets are limited in scale, realism, and environmental diversity. Most of the event-based human activity datasets either are collected under scripted laboratory settings that fail to capture the variability and spontaneity of natural human behavior~\cite{thu,paf,dvsgesture,dailyactivity,asldvs,readingdataset,dvs200,hardvs}, or from fixed viewpoints that restrict scene dynamics and yield sparse background events~\cite{thu,paf,dvsgesture,dailyactivity,asldvs,dhp19}. To facilitate dataset construction, event simulators~\cite{esim,vid2e,v2e} have been widely used to generate synthetic event streams from RGB inputs. For instance, N-EPIC-Kitchens~\cite{nepickkitchen} creates simulated events by passing EPIC-KITCHENS~\cite{epickitchen} videos through ESIM~\cite{esim}. However, prior works~\cite{synIsBad1,synIsBad2,synIsBad3} have reported notable sim-to-real gaps, with models trained on simulated events often exhibiting degraded performance when evaluated on real-world recordings. Moreover, simulated event data are inherently constrained by the frame rate and dynamic range of the source RGB videos, cannot faithfully reproduce real-world sensor noise, and fail to fully capture the distinctive properties of real event streams. Other efforts, such as UCF-Crime-DVS~\cite{ucfCrime}, capture real event streams by recording RGB videos displayed on monitors. While this setup employs real event sensors, the captured data remain constrained by the parameters of the monitor, such as refresh rate, panel type, and dynamic range, which makes them less representative of real-world dynamics~\cite{dailyactivity,ncars}. These limitations highlight the need for larger, diverse, and complex benchmark datasets that represent human activity in real-world daily-life scenarios, enabling large-scale pre-training and systematic benchmarking for neuromorphic vision beyond the autonomous driving domain.

In this paper, we introduce \dataname: a large-scale, real-world, egocentric, stereo event camera dataset. The key recording device is a pair of stereo Prophesee Gen4 event cameras, which provide high-resolution event data. To provide reliable ground truth for the event data, our setup is multi-modal: alongside an event-based stereo stream, it features an RGB-based stereo stream, an RGB-depth pair, and an Inertial Measurement Unit (IMU), as shown in Fig.~\ref{fig:teaser_figure}. All sensors are mounted on a wearable helmet to support egocentric data collection. Compared to fixed-view setups, the egocentric configuration introduces rich scene motion that generates abundant background events, which are highly informative for event-based perception tasks~\cite{nepickkitchen,EE3D-R,eventreview}.

Following established paradigms in RGB data~\cite{epickitchen,hdepic,breakfast}, we focus on cooking scenarios to capture natural human activities, as kitchens inherently involve diverse human–object interactions, rapid motion dynamics, and complex visual patterns. The dataset is collected in 13 diverse kitchen environments with 10 participants, resulting in 5.5 hours of stereo event recordings. We ask each participant to wear the data collection device and naturally perform several common cooking activities in the kitchen without any instructions, which ensures the activities are unscripted. With stereo cameras and ground-truth depth maps, \dataname enables accurate 3D perception of human motions in indoor environments. The wearable device setup of \dataname also supports research in Augmented Reality, Virtual Reality, Human-Computer Interaction, and wearable AI~\cite{ego1,ego2}. 

To demonstrate the potential and complexity of our dataset, we train preliminary baseline models on event-based action recognition, object detection, and stereo depth estimation. Our main contributions are:
\begin{itemize}
    \item We collect the first large-scale stereo event camera benchmark dataset in a real-world daily life and egocentric setting for multiple tasks. 
    \item We provide annotations of human action labels, action segments, object labels, and object bounding boxes.
    \item We supply the ground truth depth maps and calibration matrices among sensors. 
    \item We present challenges on our \dataname dataset by evaluating seven baseline models.
\end{itemize}

%% file: sec/2_relatedwork.tex
\section{Related Work}
\label{sec:relatedwork}
\begin{table*}[tp]
\centering
\caption{Comparison of our \dataname and event-based human activity datasets in terms of real-world capture, stereo event cameras,  egocentric setting, event camera resolution, unscripted recording (UR), supporting multiple tasks (MT), object of activity, number of activity classes (\#Cls), number of object bounding boxes (\#BB), number of depth maps (\#DM), and duration per recording (DR). \dataname is the only dataset recorded with stereo event cameras in the egocentric, real-world setting, and supports multiple tasks. \dataname compares favorably to existing event camera datasets on resolution and number of annotations. The average recording duration of three minutes encourages natural and unscripted behavior during data collection in \dataname. Note that we only report the dataset collected with real event cameras. 
}
\renewcommand{\arraystretch}{1.10}   
\setlength{\tabcolsep}{4.5pt}        
\begin{adjustbox}{width=\textwidth}
\begin{tabular}{@{} l c c c c c c c c c c c c @{}}
\Xhline{2\arrayrulewidth}
\multirow{2}{*}{\textbf{Dataset}} & \multirow{2}{*}{\textbf{Year}} & \textbf{Real-} & \multirow{2}{*}{\textbf{Stereo}} & \multirow{2}{*}{\textbf{Ego}} & \multirow{2}{*}{\textbf{Resolution}} & \multirow{2}{*}{\textbf{UR}} & \multirow{2}{*}{\textbf{MT}} & \multicolumn{2}{c}{\textbf{Activity}} & \multirow{2}{*}{\textbf{\#BB}} & \multirow{2}{*}{\textbf{\#DM}} & \multirow{2}{*}{\textbf{DR}} \\ \cline{9-10}
                                  &                                & \textbf{world} &                                  &                               &                                      &                              &                              & \textbf{Object}   & \textbf{\#Cls}  &                              &                              &                             \\ \hline
ASLAN-DVS~\cite{asldvs}                 & 2019    & \xmark  & \xmark  & \xmark  & $240\times 180$     & \xmark & \xmark & Action  & 432 & \xmark & \xmark & - \\
UCF-DVS~\cite{asldvs,soomro2012ucf101}  & 2019    & \xmark  & \xmark  & \xmark  & $240\times 180$     & \xmark & \xmark & Action  & 101 & \xmark & \xmark & 25s \\
HMDB-DVS~\cite{asldvs,kuehne2011hmdb}   & 2019    & \xmark  & \xmark  & \xmark  & $240\times 180$     & \xmark & \xmark & Action  & 51  & \xmark & \xmark & 19s  \\ 
UCF-Crime-DVS~\cite{ucfCrime}           & 2025    & \xmark  & \xmark  & \xmark  & $1280\times 720$    & \cmark & \xmark & Anomaly & 14  & \xmark & \xmark & avg 242s  \\ \hline
DVS-Gesture~\cite{dvsgesture}           & 2017    & \cmark  & \xmark  & \xmark  & $128\times 128$     & \xmark & \xmark & Action  & 11  & \xmark & \xmark & 6s  \\
ASL-DVS~\cite{asldvs}                   & 2019    & \cmark  & \xmark  & \xmark  & $240\times 180$     & \xmark & \xmark & Hand    & 24  & \xmark & \xmark & 0.1s  \\
PAF~\cite{paf}                          & 2019    & \cmark  & \xmark  & \xmark  & $246\times 260$     & \xmark & \xmark & Action  & 10  & \xmark & \xmark & 5s  \\ 
DailyAction~\cite{dailyactivity}        & 2021    & \cmark  & \xmark  & \xmark  & $346\times 260$     & \xmark & \xmark & Action  & 12  & \xmark & \xmark & 5s  \\
Bullying10K~\cite{bullying10k}          & 2023    & \cmark  & \xmark  & \xmark  & $346\times 260$     & \xmark & \xmark & Action  & 10  & \xmark & \xmark & 2-20s  \\
HARDVS~\cite{hardvs}                    & 2023    & \cmark  & \xmark  & \xmark  & $346\times 260$     & \xmark & \xmark & Action  & 300 & \xmark & \xmark & 5s  \\
THU$^{\text{MV-EACT}}-50$~\cite{thu}    & 2024    & \cmark  & \xmark  & \xmark  & $1280\times 800$    & \xmark & \xmark & Action  & 50  & \xmark & \xmark & 2-5s  \\
DailyDVS-200~\cite{dvs200}              & 2024    & \cmark  & \xmark  & \xmark  & $320\times 240$     & \xmark & \xmark & Action  & 200 & \xmark & \xmark & 1-20s  \\
SeAct~\cite{seact}                      & 2024    & \cmark  & \xmark  & \xmark  & $346\times 260$     & \xmark & \xmark & Action  & 58  & \xmark & \xmark & -  \\
EE3D-R~\cite{EE3D-R}                    & 2024    & \cmark  & \xmark  & \cmark  & $640\times 480$     & \xmark & \xmark & Pose    & 10  & \xmark & \xmark & -     \\  \hline 
\textbf{\dataname (ours)}               & 2026    & \cmark  & \cmark  & \cmark  & $1280\times 720$    & \cmark & \cmark  & Action & 268  & 13,482  & 297,457 & avg 180s  \\ \Xhline{2\arrayrulewidth}
\end{tabular}
\end{adjustbox}
\label{tab:comparison}
\end{table*}

We compare \dataname with representative human activity datasets captured using real event cameras in Table~\ref{tab:comparison}. Existing event camera datasets are generally acquired either by recording monitors that display RGB data~\cite{asldvs,ucfCrime,nimagenet,ncifar,nmnistcaltech}, or by directly capturing real-world scenes with event sensors~\cite{thu,dailyactivity,dvs200,hardvs,seact,dvsgesture,paf,asldvs}. However, most of them are designed for a single task. In contrast, our \dataname can be used to benchmark multiple tasks involving natural human activities within the real-world, stereo, and egocentric setting. 

\noindent\textbf{Synthetic datasets.} Due to the scarcity and cost of event cameras, the scale of event camera datasets remains far smaller than the RGB datasets. To address this, a common approach is to record RGB data displayed on monitors using an event camera, as adopted in event-based object recognition datasets N-ImageNet~\cite{nimagenet}, N-MNIST~\cite{nmnistcaltech}, N-Caltech101~\cite{nmnistcaltech}, event-based action recognition datasets ASLAN-DVS~\cite{asldvs}, UCF-DVS~\cite{asldvs}, HMDB-DVS~\cite{asldvs}, and event-based anomaly detection dataset UCF-Crime-DVS~\cite{ucfCrime}. However, it remains limited by the refresh rate and brightness range of the display setup, thus failing to fully reflect real-world dynamics.

\noindent\textbf{Real-world datasets.} THU$^{\text{MV-EACT}}$-50~\cite{thu} collects human actions with six event cameras at a high resolution of $1280\times 800$. Unfortunately, it is script-driven with 50 predefined actions. DailyDVS-200~\cite{dvs200} and HARDVS~\cite{hardvs} provide a wide range of human actions of 200 and 300 classes each, captured with low resolutions of $320\times 240$ and $346\times 260$, respectively, in diverse environments and illumination conditions. Still, their action classes are scripted. SeAct~\cite{seact}, DailyAction~\cite{dailyactivity}, DVS-Gesture~\cite{dvsgesture}, ASL-DVS~\cite{asldvs}, Bullying10K~\cite{bullying10k}, and PAF~\cite{paf} focus on action recognition, but with limited predefined action classes and low-resolution samples. EE3D-R~\cite{EE3D-R} targets 3D human pose estimation with a wearable helmet, yet it only includes 10 scripted motions captured by a single event camera.

\noindent\textbf{Complexity limitation.} Currently, popular event camera datasets~\cite {nmnistcaltech,dvsgesture,asldvs,ncars,paf} are limited in complexity. As reported in~\cite{nimagenet}, methods developed prior to 2021 already achieved over $90\%$ object recognition accuracy on~\cite{nmnistcaltech,dvsgesture,asldvs,ncars,ncifar}, suggesting that these datasets no longer pose sufficient challenges for modern models. This highlights the need for new event camera datasets that better reflect the complexity and variability of real-world environments.

To bridge the gap, our \dataname dataset provides high-resolution, multi-task, real-world data of natural human activities in a multi-modality, stereo, and egocentric setup.

%% file: sec/3_dataset.tex
\section{The \dataname Dataset}
\label{sec:dataset}

We build a data collection device with multiple sensors, capture natural human actions with multiple participants, and annotate data with human annotators.

\begin{figure*}[t]
    \centering
    \includegraphics[width=\textwidth]{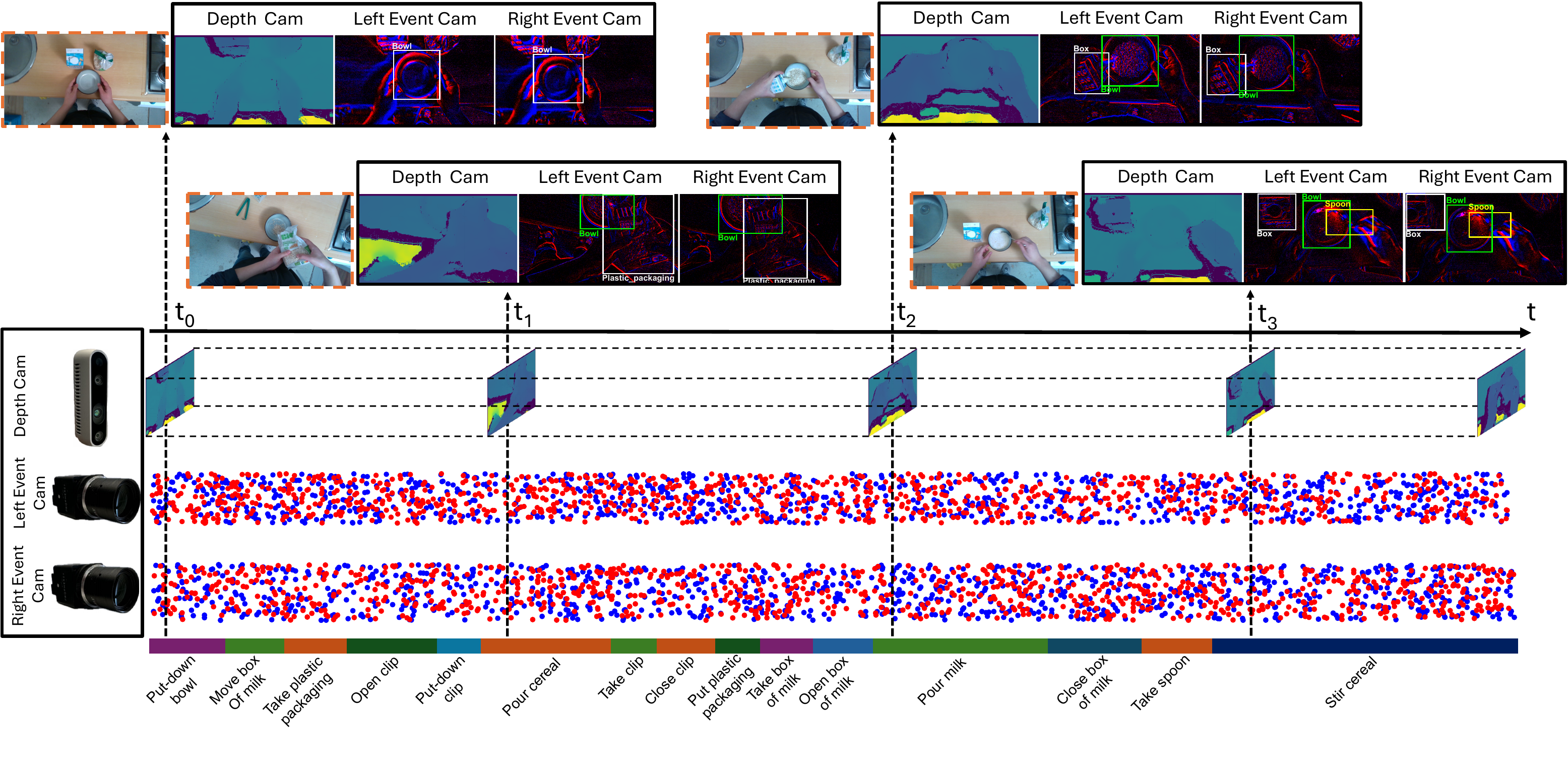}
    \caption{
    \textbf{Top:} Rectified and annotated frames from four different time points $t_0, ..., t_3$. Frames depict the ground truth depth maps and corresponding events aggregated from the left and right event cameras, along with human-annotated object bounding boxes. RGB references highlight the difficulty of annotating on events directly; \textbf{Middle:} Video streams from the depth camera, and left and right event cameras. Event colors (red, blue) show the event polarity; \textbf{Bottom:} Human-annotated action segments.}
    \label{fig:multi-task}
\end{figure*}

\begin{figure*}[t]
    \centering
    \includegraphics[width=\textwidth]{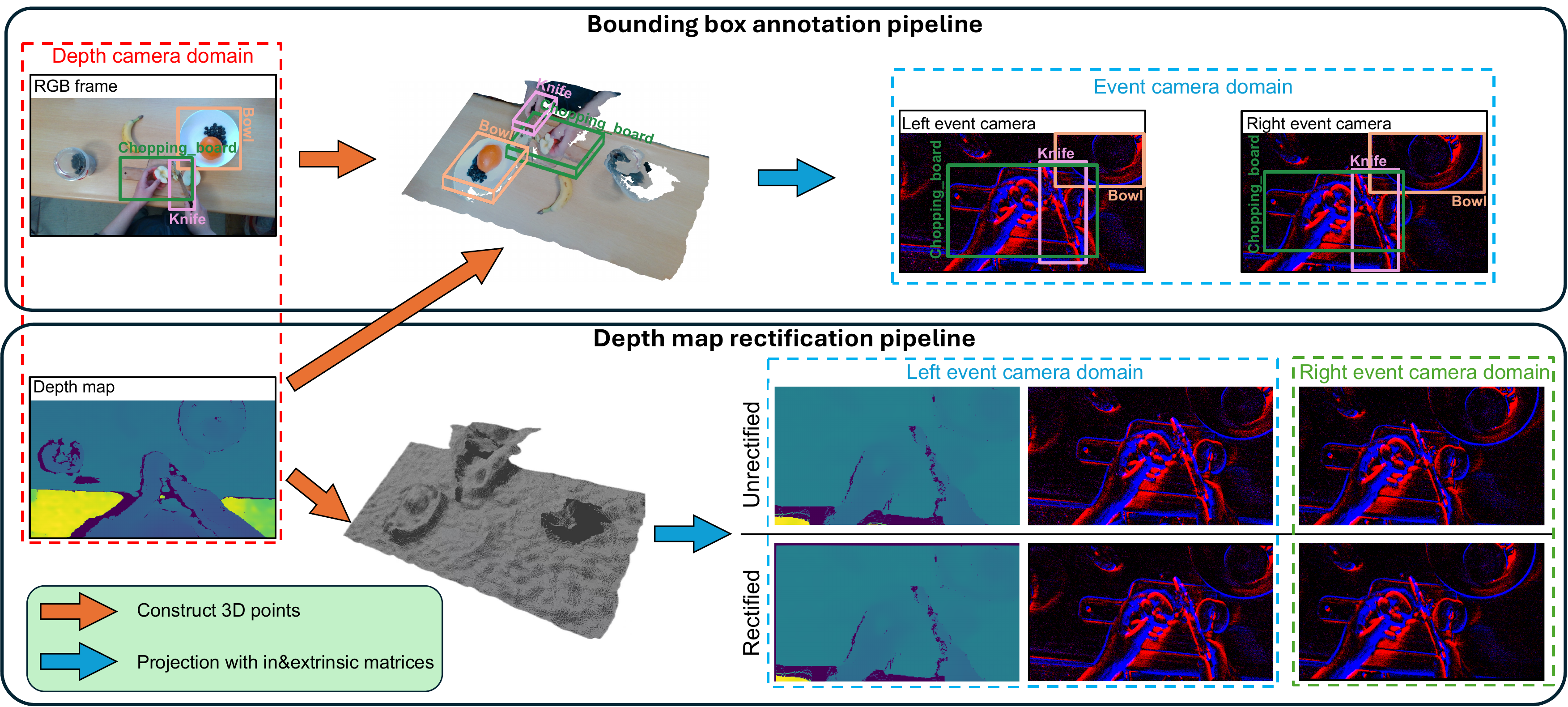}
    \caption{\textbf{Bounding box annotation pipeline:} 1. Obtain human-annotated bounding boxes on the RGB frame; 2. Construct 3D points of each bounding box based on its synchronized depth map; 3. Project 3D points to the left and right event camera using intrinsic and extrinsic matrices. \textbf{Depth map rectification pipeline:} 1. Construct 3D points for the depth map; 2. Project all 3D points to the target camera domain (left or right event camera) to get the projected depth map; 3. Rectify the projected depth with left and right event cameras.
    }
    \label{fig:bbox_depth_pipeline}
\end{figure*}

\subsection{The Data Collection Device}
Our data collection device is a wearable system mounted on a bicycle helmet, integrating multiple sensors as shown in Fig.~\ref{fig:teaser_figure}(a). Two HD Prophesee Gen4 event cameras are mounted on a 3D-printed structure to form a stereo pair. We mount two CMOS RGB cameras next to each event camera to form the RGB-EVENT unit. To acquire reliable ground truth depth measurements, an Intel RealSense depth camera is positioned above the stereo configuration, adding another RGB camera (D-RGB), depth maps, and a 6-axis IMU. (See the supplement for further device details.)

In summary, \dataname incorporates diverse multimodal data: two event streams, three RGB streams (two CMOS RGB and D-RGB), one depth stream, and one IMU stream.

\noindent\textbf{Synchronization.}
We connect all sensors to a data-recording laptop using USB-3 ports. We employ the Robot Operating System (ROS)~\cite{ros} as the framework for managing and operating all sensors in our data collection system. All sensors are synchronized with the timestamps provided in ROS.

\noindent\textbf{Calibration.} 
Due to the inherent asynchrony of event cameras, conventional calibration techniques based on image-based corner detection are not directly applicable. We adopt E2Calib~\cite{e2calib} to reconstruct event streams into grayscale frames and apply the standard OpenCV~\cite{opencv} calibration framework to perform intrinsic calibration for all cameras. For extrinsic calibration, we designate the depth camera as the reference and perform stereo calibration with the remaining cameras, respectively. To facilitate event-based stereo depth estimation, the left event camera is used as the reference for extrinsic calibration with both the right event camera and the depth camera. After calibration, both intrinsic and extrinsic re-projection errors are below two pixels for all cameras. Detailed calibration results are given in the supplement.

\subsection{Data collection}
Variability and task complexity are important assets of a benchmark dataset. \dataname includes recordings with 10 participants of different nationalities, genders, and ages in eight private kitchens and five public kitchens. The participant statistics are reported in the supplement.
As the data collection system integrates multiple sensors and is relatively heavy, we carefully considered participant comfort and safety during recording, in accordance with the HERC approval. Accordingly, we selected 13 cooking activities with moderate duration, from which participants could freely choose, which include \textit{1. Cut bread, 2. Cut cake, 3. Fry bacon, 4. Fry egg, 5. Fry pepper, 6. Make cereal bowl, 7. Make coffee, 8. Make fruit salad, 9. Make lemon water, 10. Make a sandwich, 11. Make tea, 12. Make vegetable salad,} and \textit{13. Wash dish.} 
At the start of the recording session, each participant voluntarily chooses a set of activities based on their preference. The data is recorded by the participant wearing the data collection helmet and a backpack containing the data recording laptop. We provide no instructions on how they perform these activities. We thus ensure that the collected data is natural and complex enough to reflect daily life activities in a real-world and unscripted setting.

Before the recording, participants are instructed to remove any identifiable items to ensure privacy protection. We perform sensor calibration before and after data collection.

\subsection{Annotation and Ground Truth}
\label{sec:annotation}
We provide human annotations of action segments and bounding boxes with VGG Image Annotator (VIA)~\cite{via} tool on the D-RGB video, and ground-truth depth maps from the depth camera. We project the annotations onto event data using the calibration matrices. Human annotators do not label event data directly since accurately identifying objects and actions is too challenging, as shown in Fig.~\ref{fig:multi-task} (D-RGB shown in orange dashed boxes).

\subsubsection{Action Segments}
To minimize annotator bias, we use the verb list of EPIC-KITCHENS~\cite{epickitchen} as reference and label actions as sets of $\{verb,~noun\}$. Three different, randomly assigned human annotators label action segments in three steps: the first annotator defines the action labels in time order, and the second annotator annotates the corresponding action segment. The last annotator verifies the action segments from the second step. Annotators are permitted to modify labels from the previous step to ensure accuracy and consistency. We use the precise synchronization between the D-RGB and event cameras to align the start and end timestamps for each action.

\begin{figure*}[tp]
    \centering
    \includegraphics[width=.95\textwidth]{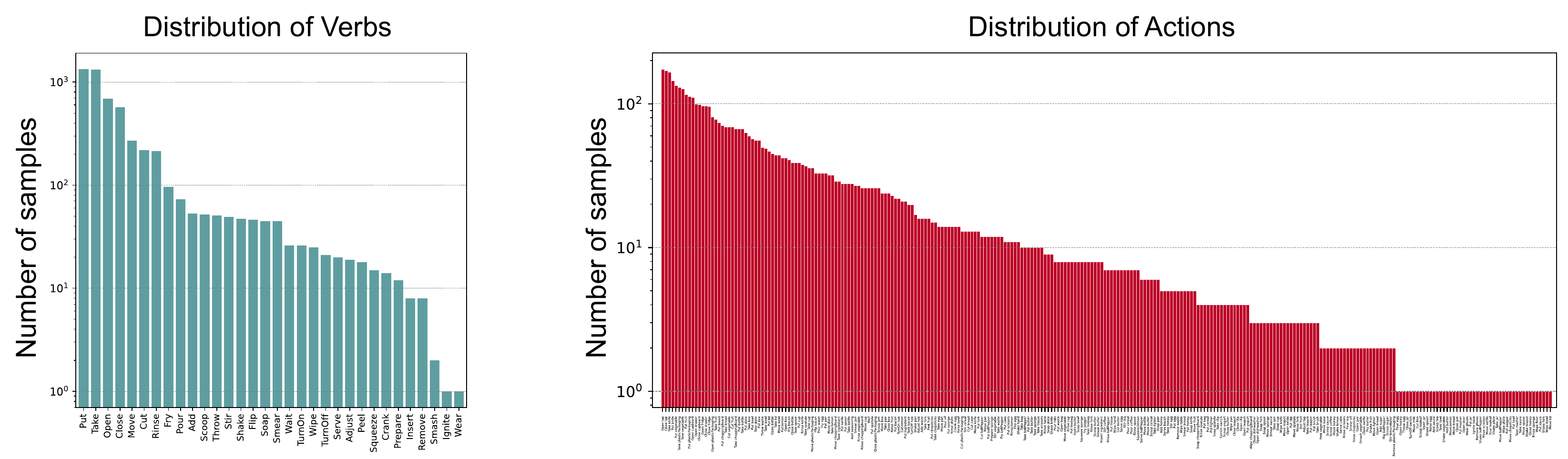}
    \caption{\textbf{Left:} Distribution of the 32 annotated action verbs in \dataname; \textbf{Right:} Distribution of the 268 annotated actions in \dataname. The two distributions demonstrate the variance of action verbs and action classes in \dataname, and the realistic long tails present further challenges, including few-shot learning and generalization. 
    \vspace{-0.5\baselineskip}
    }
    \label{fig:action_stats}
\end{figure*}

\begin{figure*}[tp]
    \centering
    \includegraphics[width=\textwidth]{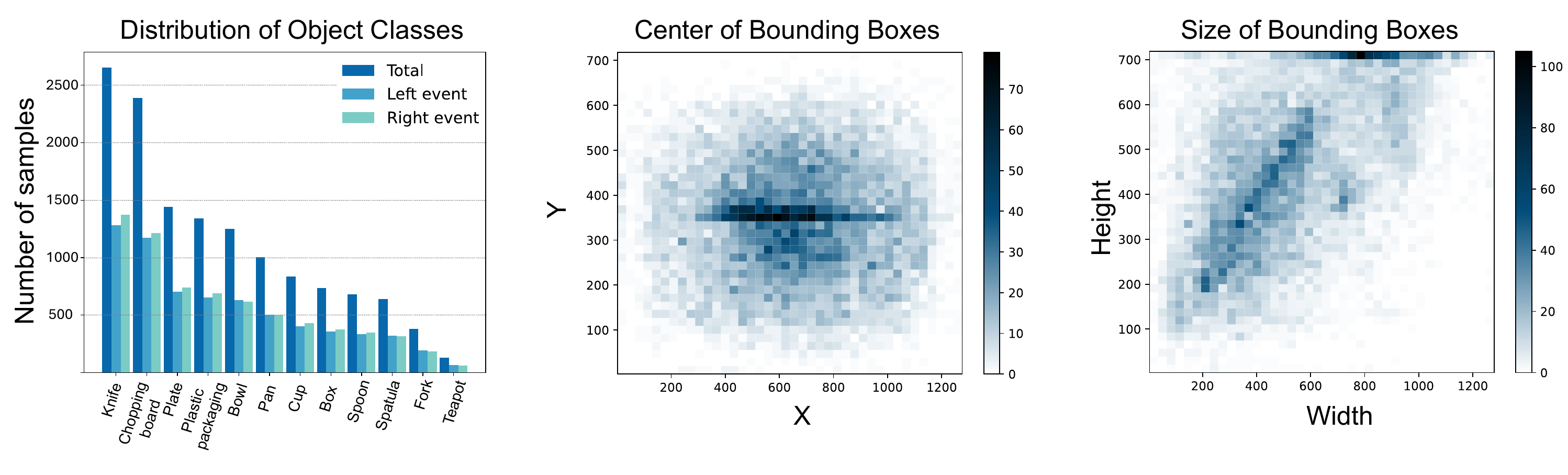}
    \caption{\textbf{Left:} Distribution of the 12 object classes in terms of number of samples in \dataname; \textbf{Middle:} Distribution of the central coordinate of all bounding boxes; \textbf{Right:} Distribution of the size of all bounding boxes. These figures demonstrate our projection pipeline is not biased, as the left and right cameras have a similar number of bounding boxes per class. Moreover, the bounding boxes are meaningfully distributed across the frames with diverse variances in size. \vspace{-1\baselineskip}
    }
    \label{fig:bbox_stats}
\end{figure*}

\subsubsection{Bounding Box Annotations}
Two human annotators are appointed to annotate the bounding boxes in each recording. For accuracy and consistency, one annotator is randomly assigned to annotate bounding boxes on the D-RGB frames, while the second annotator reviews and verifies the annotations. Given the large number and diversity of objects in kitchen environments, we focus primarily on kitchenware categories, as these objects are typically rigid and frequently encountered during cooking activities. In contrast, food items are often deformable and suffer significant appearance changes during preparation and cooking, which can introduce ambiguity in annotation~\cite{foodDeformation1,foodDeformation2}. Therefore, we restrict the object classes to the following twelve categories: \textit{1.~Bowl, 2.~Box, 3.~Chopping board, 4.~Cup, 5.~Fork, 6.~Knife, 7.~Pan, 8.~Plastic packaging, 9.~Plate, 10.~Spatula, 11.~Spoon}, and \textit{12.~Teapot}.

\textbf{Bounding box projection.}
As the D-RGB frames and depth maps are pixel-to-pixel aligned, we construct a 3D point for each pixel in the D-RGB frames (Fig.~\ref{fig:bbox_depth_pipeline}). Given a human-annotated bounding box on the D-RGB frame, we retrieve all 3D depth points inside the bounding box. Then, we project each set of 3D points to the left and right event cameras to get the projected bounding box. As refinement, we apply a smoothing process by averaging the y-axis values of all points along each width edge and the x-axis values of all points along each height edge. Bounding boxes that are out of the field of view (FoV) of the event cameras are removed. 

\subsubsection{Depth map}
\label{sec:depth}
Depth maps are captured using the Intel RealSense depth camera with an error of $<\!\!2\%$ at two meters. To enable stereo depth estimation, the ground truth needs to be rectified with the stereo system. Due to the FoV difference between the depth and event cameras, we first project the raw depth map on to the FoV of the left event camera by retrieving 3D points for all pixels, then rectify the projected depth with the event stereo. We illustrate the pipeline along with a set of rectified data in Fig.~\ref{fig:bbox_depth_pipeline}. 

\begin{figure*}[tp]
    \centering
    \includegraphics[width=\textwidth]{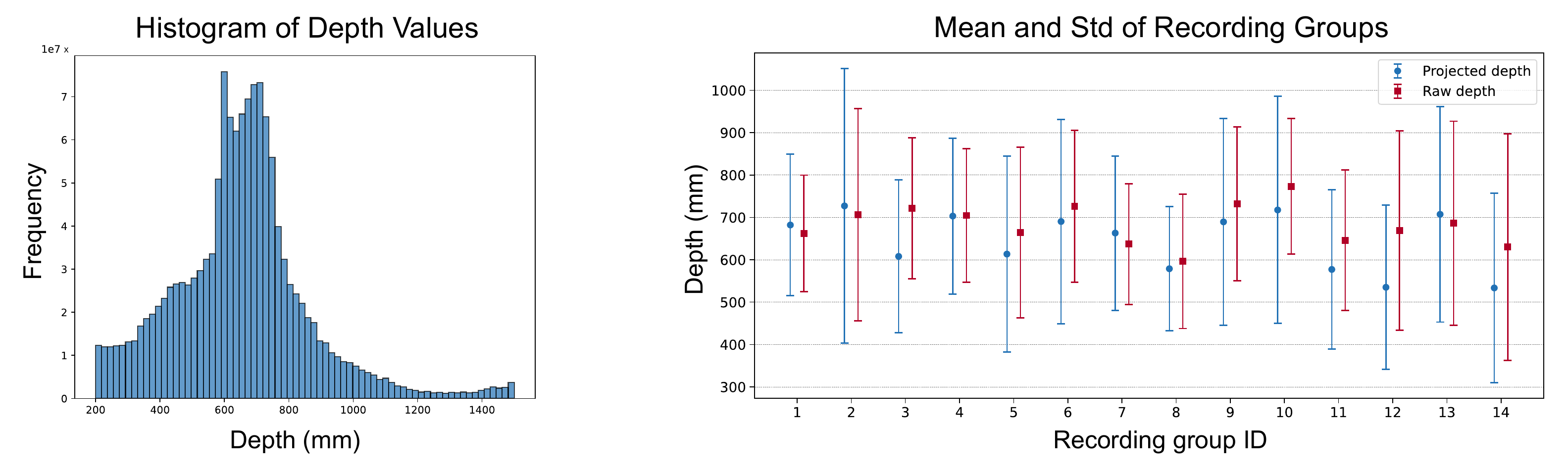} 
    \caption{\textbf{Left:} Depth histogram from all 10,800 samples is in the range from 200 mm to 1500 mm; \textbf{Right:} Mean and standard deviation of depth per session. Depth information is within an expected range for kitchen activities where objects are typically about 700~mm away.
    }
    \label{fig:depth_stats}
\end{figure*}

\subsection{Dataset Statistics}
We record 14 sessions comprising 110 sequences in 13 unique kitchen environments, totaling 5.5 hours from 10 participants, as reported in Tab.~\ref{tab:recording_stats}. With multi-sensor data, \dataname provides 5.5 hours of stereo event recordings, 5.5 hours of stereo RGB videos from CMOS cameras at 30 fps, 5.5 hours of D-RGB videos from the depth camera at 15 fps, 5.5 hours of 16-bit depth videos at 15 fps, and 5.5 hours of 6-axis IMU data at 200 fps. The average sequence duration of approximately three minutes further indicates that our recording setup preserves natural and unscripted human motions.

\begin{table*}[t]
    \centering
    \caption{Statistics of the 14 recording sessions (S1--S14). The \dataname dataset provides 14 recording sessions with a total duration of 5.5 hours. Each session consists of multiple sequences, with each sequence corresponding to a cooking activity. The average duration of each sequence is approximately three minutes.}
    \renewcommand{\arraystretch}{1.1}   
    \setlength{\tabcolsep}{4.5pt}        
    \begin{adjustbox}{width=\textwidth}
        \begin{tabular}{@{} lcccccccccccccc|c @{}}
        \Xhline{2\arrayrulewidth}
        Session ID  & S1   & S2   & S3   & S4  & S5   & S6   & S7   & S8   & S9  & S10  & S11 & S12  & S13  & S14 & All   \\ \hline
        \# Sequence  & 7    & 9    & 8    & 4   & 7    & 8    & 10   & 9    & 5   & 8    & 5   & 12   & 10   & 8   & 110   \\
        Duration (s) & 1978 & 2443 & 1333 & 537 & 1003 & 1444 & 1865 & 1746 & 491 & 1198 & 875 & 2092 & 1746 & 999 & 19750 \\ \Xhline{2\arrayrulewidth}
        \end{tabular}
    \end{adjustbox}
    \label{tab:recording_stats}
\end{table*}

\begin{table}[t]
    \centering
    \caption{Frequency ranking of common verbs in \dataname and EPIC-KITCHENS~\cite{epickitchen}. We count $\textit{`Rinse'}\ (7^{th})$ and $\textit{`Soap'}\ (16^{th})$ separately, while EPIC-KITCHENS includes the two verbs in $\textit{`Wash'}\ (3^{rd})$. 
    The similar frequency ranking between the two datasets indicates they are unscripted actions.
    }
    \renewcommand{\arraystretch}{1}   
    \setlength{\tabcolsep}{4.5pt}        
    \begin{adjustbox}{width=0.7\textwidth}
    \begin{tabular}{lccccccc}
        \Xhline{2\arrayrulewidth}
        \multirow{2}{*}{Dataset} & \multicolumn{7}{c}{Action Frequency Ranking}                                 \\ \cline{2-8} 
                                          & Put & Take & Open & Close & Move & Cut & Rinse/Wash  \\ \hline
        \dataname (ours)                        & 1$^{st}$  & 2$^{nd}$  & 3$^{rd}$ & 4$^{th}$ & 5$^{th}$  & 6$^{th}$ & 7$^{th}$    \\
        EPIC-KITCHENS~\cite{epickitchen}   & 1$^{st}$  & 2$^{nd}$  & 4$^{th}$ & 5$^{th}$ & 10$^{th}$ & 6$^{th}$ & 3$^{rd}$    \\ \Xhline{2\arrayrulewidth}
    \end{tabular}
    \end{adjustbox}
    \label{tab:verb_ranking}
\end{table}

\noindent\textbf{Action Segments.} We report the verb and action distributions in Fig.~\ref{fig:action_stats}. In total, there are 268 action classes with 32 verb classes annotated in \dataname, associated with 10,762 segments. We find that our dataset and the large-scale, unscripted egocentric EPIC-KITCHENS dataset share common high-frequency verbs (Table~\ref{tab:verb_ranking}). This implies that our recording and annotation pipeline preserves the naturality of human actions.

\noindent\textbf{Bounding Boxes}. Following the annotation pipeline described in Section~\ref{sec:annotation}, we annotate the D-RGB frames at a rate of 1 frame per 4 seconds, and project the labeled bounding boxes to the left and right event cameras. Eventually, there are 13,482 bounding boxes with 12 different classes in the event domain. We present the distribution of each object class, along with the size and central coordinates of all bounding boxes in Fig.~\ref{fig:bbox_stats}.

\noindent\textbf{Depth Map.} Our dataset includes 297,547 raw ground truth depth maps. Fig.~\ref{fig:depth_stats} shows the statistics of 10,800 raw depth maps sampled from each sequence.

We compare our \dataname with other event-based human activity datasets in Table~\ref{tab:comparison}. \dataname is the only event-based human activity dataset that is recorded naturally and unscripted, and supports multiple tasks.

%% file: sec/4_experiments.tex
\section{Baseline Models}
\label{sec:baseline}

EventKitchen’s multi-modal stereo setup enables several challenging tasks in the event domain. We establish baselines on the following event-based tasks: 1) action recognition, 2) object detection, 3) stereo depth estimation (Fig.~\ref{fig:multi-task}). For all baseline models, we sum the event representations from the left and right event cameras and use the combined event input for both training and evaluation. See the supplement for implementation details.

We perform a kitchen-level split, allocating nine kitchens to the training set and reserving the remaining four kitchens for testing. This protocol ensures that the test environments are entirely unseen during training. We report the dataset split in Table~\ref{tab:dataset_split}, which results in a rough 0.8:0.2 split ratio for training and test sets in terms of duration (79.4\% vs 20.6\%), annotated action segments (78.4\% vs 21.6\%), annotated bounding boxes (82.1\% vs 17.9\%), and number of depth maps (79.1\% vs 20.9\%). 

\begin{table*}[t]
    \centering
      \caption{The dataset split of \dataname. The split among both the data and annotations is roughly $0.8:0.2$. \#BB: number of bounding box, \#AC: number of action class, \#AS: number of action segment, \#DM: number of depth map}
    \renewcommand{\arraystretch}{1.1}   
    \setlength{\tabcolsep}{5pt}        
    \begin{adjustbox}{width=\textwidth}
        \begin{tabular}{lcccccccc}
        \Xhline{2\arrayrulewidth}
        Split & \#Kitchen & \#Session & \#Sequence & Duration (s) & \#BB & \#AC & \#AS & \#DM \\ \hline
        Train & 9   & 10        & 82         & 15683        & 11065          & 252            & 8598             & 235271      \\
        Test  & 4  & 4         & 28         & 4067         & 2417           & 168            & 2164             & 62276       \\ \Xhline{2\arrayrulewidth}
        \end{tabular}
    \end{adjustbox}

  \label{tab:dataset_split}
\end{table*}

\subsection{Action Recognition}
\noindent\textbf{Description.} We define our action recognition objective as follows: considering a sequence of events $E_{i}^{t_0:t_n}=\{e_{i}^{t_0}, e_{i}^{t_1}, ..., e_{i}^{t_n}\}$, the model should predict the corresponding action class $C_{a}=\{c_{v} + c_{n}; c_{v}\in C_{v}, c_{n}\in C_{n} \}$, where $c_{v}$ is the verb class, and $c_{n}$ is the noun class.

\noindent\textbf{Methods.} 
We implement two frame-based action recognition algorithms, Temporal Shift Module (TSM)~\cite{tsm} and Video Swin Transformer (Swin)~\cite{swin}, both of which achieve the top-2 performance on the large-scale event-based action recognition dataset HARDVS~\cite{hardvs} and DailyDVS-200~\cite{dvs200}, to evaluate their performance on \dataname. TSM is an efficient and lightweight method for video action recognition. Specifically, we use a ResNet-50~\cite{resnet} as the backbone for TSM. Swin is a powerful and flexible method that computes self-attention globally with spatial-temporal factorization. We use the largest Swin-Base as the backbone for Swin. Both TSM and Swin models are pretrained on Kinetics 400~\cite{kinetics}. 
We finetune the models on the selected 69 action classes $C_a^{69}$, which have at least five samples in the test set. Additionally, we finetune the models to recognize the corresponding verbs of $C_a^{69}$, which comprise 18 independent classes $C_v^{18}$.

\noindent\textbf{Criteria.} We use the top-1 and top-5 accuracy to evaluate the action class and verb class recognition. The accuracy is reported on the test sets of $C_a^{69}$ and $C_v^{18}$. 

\noindent\textbf{Results.} The results are presented in Table~\ref{tab:action_results}. Both methods achieve performance clearly above random chance, with Swin consistently outperforming TSM. Verb classification proves to be an easier task than action classification. However, neither method achieves high performance on either task. We further provide the per-class metrics of the 10 most frequent classes in $C_a^{69}$ and $C_v^{18}$ in Table~\ref{tab:action_verb_top10}. Both results strongly highlight the challenging nature of our \dataname.

\begin{table}[t]
\centering
\caption{Baseline results for action recognition. The low performances indicate the difficulty of \dataname.}
    \begin{adjustbox}{width=0.4\linewidth}
    \begin{tabular}{lcccc}
    \Xhline{2\arrayrulewidth}
    \multirow{2}{*}{Method\quad} & \multicolumn{2}{c}{Actions: $C_a^{69}$} & \multicolumn{2}{c}{Verbs: $C_v^{18}$} \\ \cline{2-5} 
                            & Top-1        & Top-5        & Top-1       & Top-5       \\ \hline
    TSM~\cite{tsm}                     & 19.23        & 42.26        & 38.17       & 87.20       \\
    Swin~\cite{swin}                    & 24.69        & 56.48        & 46.08       & 90.67       \\ \Xhline{2\arrayrulewidth}
    \end{tabular}
    \end{adjustbox}

\label{tab:action_results}
\end{table}

\begin{table}[t]
    \centering
    \caption{Test accuracy of the 10 most frequent action and verb classes in the training set. OT: open tap; CT: close tap; TK: take knife; PV: put vegetable; PK: put knife; TV: take vegetable; TP: take plastic packaging; OD: open drawer; OC: open cupboard; CD: close drawer.}
    \renewcommand{\arraystretch}{1.05}   
    \setlength{\tabcolsep}{2.5pt}        
    \begin{adjustbox}{width=\linewidth}
        \begin{tabular}{lcccccccccc|cccccccccc}
        \Xhline{2\arrayrulewidth}
        \multirow{2}{*}{Method}         & \multicolumn{10}{c|}{\textbf{10 most frequent (in train set) actions}} & \multicolumn{10}{c}{\textbf{10 most frequent (in train set) verb}}\\ \cline{2-21} 
                 & OT   & CT   & TK   & PV   & PK   & TV  & TP   & OD   & OC   & CD   & Put   & Take   & Open   & Close   & Move   & Rinse  & Cut   & Fry   & Pour   & Throw\\ \hline
        TSM~\cite{tsm}      & 20.0 & 69.6 & 24.3 & 17.6 & 23.4 & 0.0 & 32.7 & 25.0 & 34.8 & 15.0 & 54.8 & 42.3 & 42.2 & 37.7 & 5.0 & 20.6 & 44.6 & 13.9 & 4.5 & 21.4\\ 
        Swin~\cite{swin}     & 50.0 & 34.8 & 28.4 & 23.5 & 31.3 & 0.0 & 32.7 & 43.8 & 21.7 & 5.0  & 66.4 & 41.3 & 48.5 & 22.7 & 11.7 & 8.8 & 87.8 & 16.7 & 54.5 & 0.0\\
        \Xhline{2\arrayrulewidth}
        \end{tabular}
    \end{adjustbox}
    \label{tab:action_verb_top10}
\end{table}

\subsection{Object Detection}
\noindent\textbf{Description.} We define the object detection challenge as follows: within a time window of $\Delta t$, given a sequence of events $E_{i}^{t_0:t_0+\Delta t}=\{e_{i}^{t_0}, e_{i}^{t_1}, ..., e_{i}^{t_0+\Delta t}\}$, we aim to localize and classify objects of interest into 12 classes. 

\noindent\textbf{Methods.} We evaluate the state-of-the-art frame-based object detection algorithm YOLOv10~\cite{yolov10}, event-based object detection algorithm Recurrent Vision Transformers (RVT)~\cite{rvt} and EvRT-DETR~\cite{torbunov2025evrt} on \dataname. We implement the largest YOLOv10-x model pre-trained on MS-COCO~\cite{coco} and the RVT-Base model pretrained on 1 Mpx~\cite{1mpx}. We follow the EvRT-DETR pipeline: we first train the RT-DETR-B frame detector initialized from a 1 Mpx pretrained model; we then train the EvRT-DETR-B model by initializing it from the trained RT-DETR-B model.

\noindent\textbf{Criteria.} For evaluation we report the average precision $AP$~\cite{coco} with IoU threshold between 0.5 and 0.95 with steps of 0.05, $AP_{50}$ with IoU threshold at 0.5, and $AP_{05}$ with IoU threshold at 0.05.

\noindent\textbf{Results.} Table~\ref{tab:object_results} shows the baseline results.  YOLOv10 achieves an AP of 16.2\% on \dataname, while it achieves an AP of 54.4\% on the COCO dataset. RVT achieves an AP of 7.6\% on \dataname, while it achieves an AP of 47.4\% on the 1 Mpx dataset. EvRT-DETR achieves an AP of 8.5\% on \dataname, while it achieves an AP of 50.1\% on the 1 Mpx dataset. Even for $AP_{05}$, YOLOv10 reaches only 38.1\%, and both RVT and EvRT-DETR reach only 22.7\%. Although RVT and EvRT-DETR are designed for event-based object detection, they still perform below YOLOv10 on \dataname due to their specialization for high-frequency (non-human) annotations of 1Mpx automotive datasets. And we observe that the detection rates of both baselines are especially low for the classes `fork' and `spoon', as the two classes are smaller in size, often occluded by hands and food, and involve fast actions such as `stir'. These findings demonstrate the object detection complexity in our dataset

\begin{table*}[t]
    \centering
    \caption{Baseline results for object detection. The column `All' represents the mean average precision (mAP) over the 12 object classes. CB: chopping board; PP: Plastic packaging.}
    \renewcommand{\arraystretch}{1.1}   
    \setlength{\tabcolsep}{4pt}        
    \begin{adjustbox}{width=\textwidth}
        \begin{tabular}{ll|cccccccccccc|c}
        \Xhline{2\arrayrulewidth}
                     Method & Metric  & Bowl & Box  & CB   & Cup  & Fork & Knife & Pan  & PP   & Plate & Spatula & Spoon & Teapot & All  \\ \hline
        \multirow{3}{*}{{YOLOv10~\cite{yolov10}}} & $AP$      & 26.0 & 6.3  & 32.2 & 18.1 & 1.1  & 8.2   & 36.5 & 10.3 & 26.8  & 14.9    & 1.0   & 13.4   & 16.2 \\
                                             & $AP_{50}$ & 44.6 & 13.3 & 56.1 & 37.4 & 1.9  & 19.3  & 47.0 & 19.4 & 52.5  & 38.2    & 5.2   & 23.4   & 29.9 \\
                                             & $AP_{05}$ & 49.3 & 18.3 & 66.4 & 54.5 & 2.2  & 42.1  & 49.4 & 26.2 & 55.5  & 51.0    & 16.0  & 25.9   & 38.1 \\ \hline
        \multirow{3}{*}{{RVT~\cite{rvt}}} & $AP$      & 18.3 & 1.8  & 14.5 & 8.3  & 0    & 0.2   & 25.0 & 3.8  & 15.9  & 0.4     & 0.1   & 3.3    & 7.6 \\
                                             & $AP_{50}$ & 30.6 & 4.7  & 27.1 & 25.2 & 0    & 0.6   & 49.8 & 13.1 & 36.5  & 1.6     & 0.5   & 4.0    & 16.1 \\
                                             & $AP_{05}$ & 33.6 & 7.8  & 31.3 & 34.0 & 0    & 10.4  & 59.6 & 20.9 & 43.7  & 6.9     & 17.6  & 6.1    & 22.7 \\ \hline 
        \multirow{3}{*}{{EvRT-DETR~\cite{torbunov2025evrt}}} & $AP$                     & 20.5 & 1.0  & 20.4 & 6.4  & 0    & 3.2   & 27.5 & 6.0  & 14.2  & 1.5   & 0.2   & 0.5    & 8.5 \\
                                             & $AP_{50}$ & 37.7 & 2.0  & 41.0 & 18.1 & 0    & 8.0   & 37.9 & 11.8 & 35.0  & 6.1   & 1.0   & 1.4    & 16.7 \\
                                             & $AP_{05}$ & 40.2 & 6.9  & 51.1 & 27.5 & 0.2  & 25.3  & 40.0 & 18.0 & 37.4  & 16.9  & 6.5   & 2.5    & 22.7 \\                                              
                                             \Xhline{2\arrayrulewidth}
        \end{tabular}
    \end{adjustbox}
    \label{tab:object_results}
\end{table*}

\subsection{Stereo Depth Estimation}
\noindent\textbf{Description.} We define the stereo depth estimation challenge as follows: given event sequences from the left event camera $E_{l}^{t_0:t_0+\Delta t}=%
\left\{e_l^t\right\}_{t=t_0}^{t_0+\Delta t}%
$ and the right event camera 
$E_{r}^{t_0:t_0+\Delta t}=%
\left\{e_r^t\right\}_{t=t_0}^{t_0+\Delta t}%
$ that within a time window $\Delta t$, we estimate the depth map $D_{l}^{t_0+\Delta t}$ on the left event camera domain. Note that it can also be used to estimate the depth map of the right event camera.

\noindent\textbf{Methods.} We train a state-of-the-art event-based stereo depth estimation algorithm SE-CFF~\cite{secff} from scratch on \dataname. We evaluate the model in `Event-only' setting without using the RGB images. For event representation, the method uses Stacking by Number (SBN)~\cite{secff,mostafavi2021e2sri,wang2019event}. Additionally, we test an RGB stereo foundation model for zero-shot stereo matching~\cite{wen2025stereo}, where we first reconstruct left and right events to grayscale images using E2VID~\cite{Rebecq19cvpr}, then feed the reconstructed images to the foundation model. For both methods, we only train and test in the depth range from 200 mm to 1500 mm to filter invalid depth values.

\noindent\textbf{Criteria.}  We report the Root Mean Square Error (RMSE) and Mean Absolute Error (MAE) of the two methods, averaged over depth images. As a reference for estimation performance, we also include the average standard deviation (STD) and mean absolute deviation (MAD) of depth values in the ground truth. All metrics are measured in millimeters.

\noindent\textbf{Results.} Table~\ref{tab:depth_comparison} summarizes results for both methods. For SE-CFF, we report performance using the default configuration with a stack size of 5 million events at two sampling rates (1~Hz and 3~Hz), and a 15 million event stack size at 1~Hz. We find that increasing the number of events has a larger impact on error than increasing the sampling rate, which we attribute to the high spatial resolution of \dataname. Comparing the test estimation errors with the ground truth deviations (STD and MAD), the depth estimation algorithm is effective. By contrast, the higher error rates of the RGB FoundationStereo model highlight the need for adaptation strategies to better convert event data into frame-based inputs for foundation models. In both methods, the magnitude of the errors on our \dataname indicates the overall difficulty of the dataset.

\begin{table}[t]
    \centering
    \caption{%
  Depth estimation performance (in mm). Lower RMSE and MAE indicate better accuracy. 
  Errors smaller than STD and MAD in SE-CFF suggest the algorithm is effective, which is not the case for the direct utilization of FoundationStereo model. $a$M@$b$Hz denotes a maximum stack size of $a$ million events at a sampling rate of $b$~Hz.
}
    \begin{adjustbox}{width=0.6\textwidth}
    \begin{tabular}{lccc}
        \Xhline{2\arrayrulewidth}
        \textbf{Method} & \textbf{Training Setting} & \textbf{RMSE} $\downarrow$ & \textbf{MAE} $\downarrow$ \\ 
        \Xhline{1\arrayrulewidth}
        \multirow{3}{*}{SE-CFF~\cite{secff}} 
            & 5M@3Hz     & 88.19  & 59.32 \\
            & 5M@1Hz     & 88.21  & 57.38 \\
            & 15M@1Hz    & \textbf{84.91}  & \textbf{54.84} \\ 
        \hline
        FoundationStereo~\cite{wen2025stereo} & ---        & 155.06 & 123.85 \\
        \hline
        Ground Truth & ---   & {STD: 106.26} & {MAD: 82.00} \\
        \Xhline{2\arrayrulewidth}
    \end{tabular}
    \end{adjustbox}
    \vspace{-1\baselineskip}
\label{tab:depth_comparison}
\end{table}

\subsection{Discussion}
Our baseline models across three distinct tasks highlight the novel challenges and complexity of \dataname. Given that \dataname is naturally collected in 13 diverse kitchens, the data inhabits complex, variable, and dynamic background and foreground information. Thus, different scenes may map to the same action class: turning on the electrical vs. gas hobs, opening the single-lever vs. rotated handle taps, opening embedded vs. stand-alone fridges, and so on. Different participants have different cooking styles, which bring various human actions to \dataname. Take the `fry egg' activity as an example: 2 participants use chopsticks to fry eggs, 5 participants use a spatula to fry eggs, and 1 participant does not use a tool to fry eggs. These challenges demonstrate that \dataname is more complex than fixed viewpoint, studio environments, and scripted action setups, encouraging the community to develop more robust, generalizable solutions.

\noindent\textbf{Limitations.} As a naturally collected real-world dataset, we observe the long-tail distributions in both object classes and action classes of \dataname in Fig.~\ref{fig:action_stats}. Class imbalance remains a significant challenge for baseline models in object detection and action recognition. \dataname encourages future research to address this limitation.
Besides, although \dataname provides valuable human-annotated labels, the quantity remains limited compared to the 1Mpx datasets with 25 million non-human annotations. Expanding the volume of annotations presents an important direction for future development.

%% file: sec/5_conclusion.tex
\section{Conclusion}
\label{sec:conclusion}

In this work, we present a large-scale, stereo event camera dataset, \textbf{\dataname}, to support human activity-related research for the event community. Unlike existing event-based datasets, \dataname captures natural human actions across 13 diverse kitchen environments, offering 14 sessions of rich multi-sensor recordings from 10 participants. It provides 5.5 hours of stereo event recordings with synchronized RGB, depth, and IMU data, along with detailed human annotations of 10,762 action segments and 13,482 bounding boxes. Our baseline results demonstrate that \dataname offers complex new challenges for event-based 1) action recognition, with natural actions, dynamic backgrounds, and varying environments, 2) object detection, from an egocentric perspective and small, fast-moving objects, and 3) stereo depth estimation in an indoor, near-field setting. By capturing natural, real-world human activities, \dataname establishes a challenging benchmark for neuromorphic vision beyond autonomous driving. Overall, our results showcase the potential of \dataname in advancing research on daily human activities using event cameras.

%% file: sec/suppl.tex
\title{Cooking beyond Frames: A Stereo Event Camera Dataset in the Kitchen 
\\
Supplementary Material} 

\titlerunning{Cooking beyond Frames: A Stereo Event Camera Dataset in the Kitchen}

\author{Chengming Feng\inst{1}\orcidlink{0009-0008-0314-4735} \and
Hesam Araghi\inst{1}\orcidlink{0000-0002-4539-4408} \and
Liming Zheng\inst{1}\orcidlink{0000-0002-7544-3020} \and
Julien Dupeyroux\inst{2}\orcidlink{0000-0002-7414-5021} \and
Xucong Zhang\inst{1}\orcidlink{0000-0002-8368-3542} \and
Jan van Gemert\inst{1}\orcidlink{0000-0002-3913-2786} \and
Nergis Tömen\inst{1}\orcidlink{0000-0003-3916-1859}
}

\authorrunning{C.Feng et al.}

\institute{Delft University of Technology, The Netherlands \\ \email{\{c.feng-1, n.tomen\}}@tudelft.nl \and
STMicroelectronics, France
}
\maketitle
\begin{figure*}[h]
    \centering
  \includegraphics[width=\textwidth]{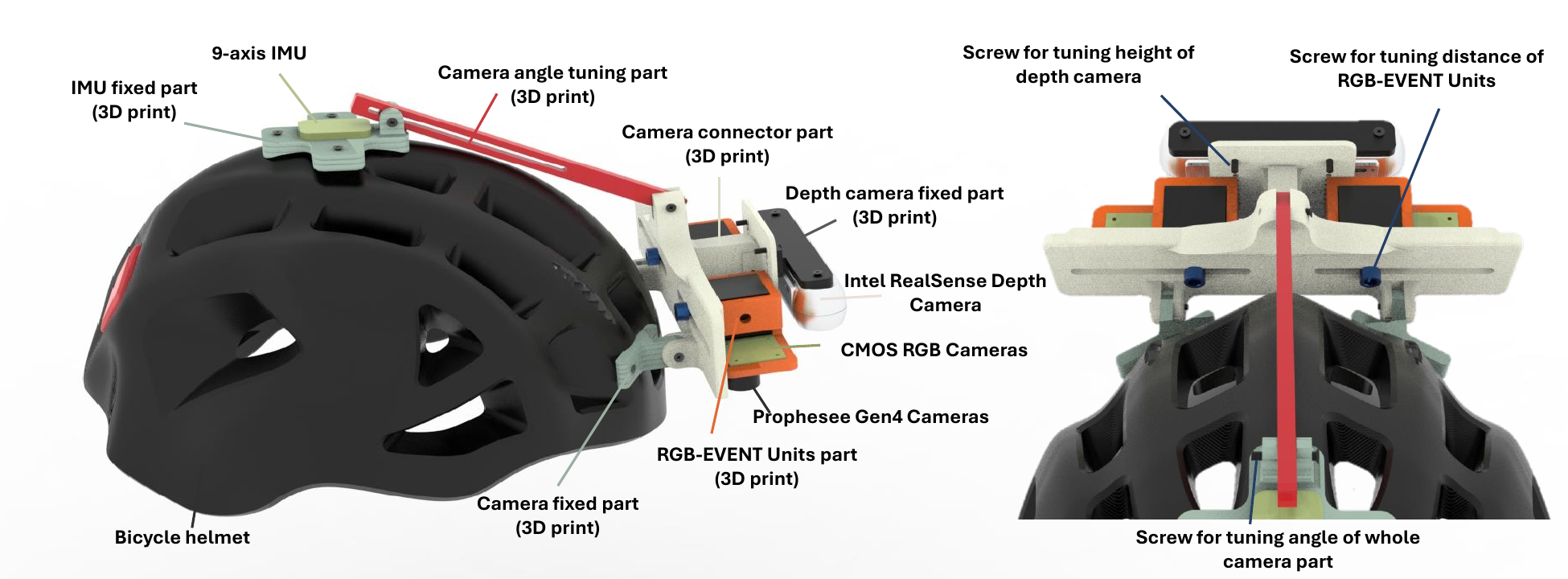}
    \caption{\textbf{Detailed illustration of our data collection device.} The device is composed of a bike helmet and a well-designed 3D structure, which ensures precise angle and distance adjustments with quick and easy installation and removal. This design facilitates flexible fitting to different participants. The 3D structure loads two Prophesee GEN4 event cameras, two CMOS RGB cameras, one Intel RealSense camera, and one 9-axis IMU, which enables multi-modal data collection. The IMU is not included in the main paper because its calibration relative to the camera module has not been fully validated.
    }
    \label{fig:design}
\end{figure*}

\section{Data Collection Device Design}
Considering the irregular shape of bike helmets and variations in head shapes among users, we designed the data collection device to allow fine-tuned adjustments in both angle and distance while ensuring quick installation and removal (see  Fig.~\ref{fig:design}). The device consists of an IMU module and a camera module. The IMU module is mounted on a 3D-printed base, which is secured to the top of the helmet using four M3 screws and adhesive. Although the rig physically includes this IMU, we exclude it from the main paper because its calibration relative to the camera module has not been fully validated.

The camera module includes three types of cameras. Two RGB-EVENT Units, each consisting of a CMOS RGB camera and a full HD Prohesee Gen4 event camera, are symmetrically mounted at the front of the helmet. These cameras are connected via a 3D-printed bracket, ensuring that their CMOS sensors remain in the same horizontal plane. The relative distance between the two RGB-EVENT Units can be adjusted using an M6 screw at the rear. Additionally, an Intel RealSense D435i (or D455) Depth Camera is installed at the frontmost position of the device, with its height finely adjustable via a 3D-printed Depth Camera Fixed Part. The IMU and camera modules are connected through a rotatable, slotted 3D-printed linkage, allowing precise control over the camera's shooting angle via a screw mechanism. This modular and adjustable design ensures compatibility with various helmet types while maintaining stable and flexible data collection.

\section{Statistics of \dataname}
\subsection{Recording}
\dataname is collected in 13 diverse kitchens, including 8 private kitchens and 5 public kitchens. A total of 10 adult participants took part in the recordings, with 7 participants identifying as men and 3 participants identifying as women. Overall, the participants are from 4 different nationalities.

\subsection{Data Privacy}
All participant data is anonymized according to the applicable rules and regulations. A Human Research Ethics Committee (HREC) approval was obtained before starting the recordings. Since some of the recordings took place in participants' private kitchens, they were instructed to remove any personally identifiable items in advance. All participants signed consent forms before conducting the recordings. The recordings, as well as the cleaned, curated and annotated dataset will be stored and made publicly available in accordance with the local and international regulations.

\subsection{Calibration}
We report the stereo calibration results in Table~\ref{tab:calibration_result}. We perform the stereo calibration among sensors to obtain the extrinsic matrices. For each stereo pair, the projection error is lower than 2 pixels.

\begin{table}[h]
  \centering
  \caption{The projection error of each stereo pair in units of pixels. Our data collection device is well-calibrated with the reprojection error lower than 2 pixels. D435i: RealSense D435i camera; D455: RealSense D455 camera; LEvent: left event camera; REvent: right event camera.}
  \begin{adjustbox}{width=.5\linewidth}
    \begin{tabular}{cccc}
    \Xhline{2\arrayrulewidth}
    \multicolumn{2}{c}{Stereo Pairs}                                 & \multicolumn{2}{c}{Reprojection Error} \\ \hline
    Camera A                                & Camera B               & Camera A           & Camera B          \\ \hline
    \multirow{2}{*}{D435i} & LEvent      & 1.068              & 0.955             \\ \cline{2-2}
                                            & REvent     & 0.944              & 0.956             \\ \hline
    \multirow{2}{*}{D455}  & LEvent      & 0.681              & 1.022             \\ \cline{2-2}
                                            & REvent     & 0.425              & 0.562             \\ \hline
    \multirow{3}{*}{LEvent}      & D435 & 0.955              & 1.068             \\ \cline{2-2}
                                            & D455  & 1.022              & 0.681             \\ \cline{2-2}
                                            & REvent     & 1.031              & 0.680             \\ \Xhline{2\arrayrulewidth}
    \end{tabular}
  \end{adjustbox}

  \label{tab:calibration_result}
\end{table}

\section{Baseline Models}
We provide the implementation details of each baseline model below. All baseline models: TSM~\cite{tsm}, Swin~\cite{swin}, YOLOv10~\cite{yolov10}, RVT~\cite{rvt}, EvRT-DETR~\cite{torbunov2025evrt}, SE-CFF~\cite{secff}, FundationStereo~\cite{wen2025stereo}, and corresponding data preprocessing with E2VID~\cite{Rebecq19cvpr}, are publicly available, and we follow their official GitHub repositories to conduct the experiments. A random seed of 42 is set for all baseline models to ensure reproducibility. 
\subsection{Action Recognition}
Following \cite{nepickkitchen}, we create a voxel grid~\cite{voxel}, implemented using Tonic~\cite{tonic}, with three channels and 100~ms time bins to train the TSM~\cite{swin}. A uniform sampling of five voxel grids per action segment is used to create the input for TSM. We use the SGD~\cite{sgd} optimizer with an initial learning rate ${\alpha=0.001}$, a momentum $\mu=0.9$, and a weight decay $\lambda=10^{-7}$ for fine-tuning the ResNet-50~\cite{resnet} pretrained on Kinetics-400~\cite{kinetics}. A step decay of 0.1 per 20 epochs is utilized to decay the learning rate. We train the network for 60 epochs with a batch size of four on one NVIDIA A40 GPU. We implement a multi-scale crop and a horizontal flip with a probability of 0.5 as data augmentation following \cite{nepickkitchen}.

We use the event frames method~\cite{eventframe} with a time window of $1/30$~s to generate input frames, and normalize frames to the scale $0-1$ to train Swin~\cite{swin}. Per action segment, a uniform sampling of 16 frames is implemented. We use the AdamW~\cite{adamw} optimizer with an initial learning rate $\alpha=0.001$, momentum $\beta_1=0.9, \:\beta_2=0.999$, and a weight decay $\lambda=0.05$ for fine-tuning the Swin-B pretrained on Kinetics-400. A Cosine Annealing schedule with a minimum learning rate at zero and a linear warmup with 2.5 epochs is applied. We train the network for 60 epochs with a batch size of four on one NVIDIA A40 GPU. We implement a horizontal flip with a probability of 0.5 as data augmentation.

For evaluation, we choose the standard top-1 and top-5 accuracy~\cite{tsm,swin,dvs200,hardvs,epickitchen,nepickkitchen} to test the performance of baselines.

\subsection{Object Detection}
To train YOLOv10, we use a time window of $1/30$~s to accumulate events into the 2-channel frame representations with the positive and negative polarities. All event frames are normalized to $0-255$. We use the SGD \cite{sgd} optimizer with an initial learning rate $\alpha=0.01$, momentum $\mu=0.937$, weight decay $\lambda=5\times10^{-4}$, and a linear scheduler following the implementation of~\cite{yolov10}. All experiments are conducted with 100 epochs and a batch size of eight on a single NVIDIA A40 GPU. As augmentation, we use flipping and scaling with a probability of 0.5 each, and event erasing with a probability of 0.4.

We follow~\cite{rvt} to create the input representation, namely \textit{Stacked Histogram}, of 10 bins within a time window of $50$~ms. We use the Adam~\cite{adam} optimizer with an initial learning rate $\alpha=0.0006$,  weight decay $\lambda=5\times10^{-4}$, and a OneCycle learning rate schedule~\cite{onecycle}. We train the model for 60 epochs and a batch size of 16 on a single NVIDIA A40 GPU. Flipping with a probability of 0.5 is applied as the data augmentation. 

For EvRT-DETR, we use the same event representation as RVT, namely a \textit{Stacked Histogram} with 10 temporal bins over a $50$~ms window. We train EvRT-DETR in two stages following the default pipeline. First, we train a frame-based RT-DETR detector with a ResNet-50 backbone, initialized from a 1 Mpx-pretrained RT-DETR checkpoint. This stage uses AdamW with learning rate $\alpha=10^{-4}$, detection-head learning rate $10^{-3}$, weight decay $0$, batch size 48, and focal classification loss. We then initialize the video-based EvRT-DETR from the best frame checkpoint. The video stage is trained with clips of length 8 and batch size 12 using AdamW with learning rate $\alpha=0.0006$, weight decay $0$, and a OneCycle learning rate schedule. Horizontal flipping is used in both stages, while the video stage additionally applies geometric augmentation with probability $0.6$ and random erasing with probability $0.4$. All experiments are run on a single NVIDIA A40 GPU.

We use the average precision $AP$~\cite{coco} with IoU threshold between 0.5 and 0.95 with steps of 0.05, and $AP_{50}$ with IoU threshold at 0.5 to evaluate the object detection models, as they are the standard metric for object detection~\cite{yolov10,rvt,coco}.

\subsection{Stereo Depth Estimation}
We train the SE-CFF~\cite{secff} model from scratch for 100 epochs with a batch size of 4 on two NVIDIA A40 GPUs with 48~GB of memory each. For event representation, the method employs Stacking by Number (SBN)~\cite{secff,mostafavi2021e2sri,wang2019event} which concatenates 10 sequences of multi-scale stacks created with a variable number of events per stack. In addition to the default largest stack size of 5 million events, we also test a larger size of 15 million events, considering the increased event count at higher resolutions. We train with two sampling rates, 1~Hz and 3~Hz, and test at 3~Hz. During training, we apply random cropping augmentation, reducing the input resolution to 90\% ($1152\times648$). 
The remaining hyperparameters follow the original SE-CFF paper. Specifically, the network weights are initialized with random values. For optimization, we use the Adam optimizer~\cite{adam} with a learning rate of $5 \times 10^{-4}$ and a weight decay of $10^{-4}$. 
The learning rate is scheduled using cosine annealing with a warmup period of 3 epochs. 
For the concentration network, the number of base channels is set to 32, and for the disparity estimation network, the maximum disparity is set to 192.

For testing FoundationStereo~\cite{wen2025stereo}, a preprocessing step is required. We first reconstruct events from the left and right event cameras into grayscale images using E2VID~\cite{Rebecq19cvpr} with a time window of $1/30$~s on an NVIDIA RTX 4090 GPU. The reconstructed left and right grayscale images are then synchronized with the depth ground truth using their timestamps. Next, both the grayscale images and the depth map are rectified using the calibration matrices. For depth inference, we use the pretrained ViT-Large model provided by the authors, and perform inference on an NVIDIA A40 GPU with 48~GB of memory. We neither apply hierarchical inference nor downsample the input images. During inference, the number of flow-field updates in the forward pass is set to 32. As in the original paper, three levels of GRU blocks are used for hidden state updates in each iteration.

We also experimented with brightness gamma correction (with $\gamma = 1.5$) to reduce the domain gap between the reconstructed images and the input format expected by the FoundationStereo model, aiming to improve adaptation to event-based inputs. 
However, as shown in Table~\ref{tab:gamma_correction}, the error rates were higher with gamma correction, suggesting that this preprocessing step is not suitable in our setting.

\begin{table}[t]
\centering
\caption{Effect of gamma correction on reconstructed left and right images (from events) when tested with the FoundationStereo model.}
\begin{tabular}{lcc}
\Xhline{2\arrayrulewidth}
Method & RMSE & MAE \\ \hline
Without gamma correction & 155.06 & 123.85 \\
With gamma correction & 159.87 & 129.23 \\ \Xhline{2\arrayrulewidth}
\end{tabular}
\label{tab:gamma_correction}
\end{table}

To evaluate the depth estimation models, we choose the Root Mean Squared Error (RMSE) and Mean Absolute Error (MAE) as they are the most commonly used metrics in the field. 

%
%